\documentclass{article}
\usepackage[preprint]{neurips_2021}
\usepackage{trackchanges}
\usepackage{amssymb}
\usepackage[utf8]{inputenc} 
\usepackage[T1]{fontenc}    
\usepackage{hyperref}       
\usepackage{url}            
\usepackage{booktabs}       
\usepackage{amsfonts}       
\usepackage{nicefrac}       
\usepackage{microtype}      
\usepackage{xcolor}         
\usepackage{multicol}
\usepackage{multirow}
\usepackage{wrapfig}
\usepackage{amsmath,amsthm,mathtools,amsfonts,amssymb}
\usepackage{graphicx,subcaption}
\usepackage{algorithm, algorithmic}

\title{LordNet: An Efficient Neural Network for Learning to Solve Parametric Partial Differential Equations without Simulated Data}

\author{%
Xinquan Huang$^{1}$\thanks{Equal contribution} \ \thanks{Work done during an internship at Microsoft Research AI4Science} \ , Wenlei Shi$^{2}$\footnotemark[1] \ , Xiaotian Gao$^{2}$\footnotemark[1]\\
\textbf{Xinran Wei$^{2}$, Jia Zhang$^{2}$, Jiang Bian$^{2}$, Mao Yang$^{2}$, Tie-Yan Liu$^{2}$}\\
$^1$King Abdullah University of Science and Technology \quad $^2$Microsoft Research AI4Science \\
\texttt{\{xinquan.huang\}@kaust.edu.sa} \\
\texttt{
\{wenlei.shi,xiaotian.gao,weixinran,jia.zhang\}@microsoft.com}\\
\texttt{\{jiang.bian,maoyang,tie-yan.liu\}@microsoft.com} \\
}

\begin{document}
\maketitle
\begin{abstract}
Neural operators, as a powerful approximation to the non-linear operators between infinite-dimensional function spaces, have proved to be promising in accelerating the solution of partial differential equations (PDE).
However, it requires a large amount of simulated data, which can be costly to collect. 
This can be avoided by learning physics from the physics-constrained loss, which we refer to it as mean squared residual (MSR) loss constructed by the discretized PDE. 
We investigate the physical information in the MSR loss, which we called long-range entanglements, and identify the challenge that the neural network requires the capacity to model the long-range entanglements in the spatial domain of the PDE, whose patterns vary in different PDEs.
To tackle the challenge, we propose LordNet, a tunable and efficient neural network for modeling various entanglements. 
Inspired by the traditional solvers, LordNet models the long-range entanglements with a series of matrix multiplications, which can be seen as the low-rank approximation to the general fully-connected layers and extracts the dominant pattern with reduced computational cost.
The experiments on solving Poisson's equation and (2D and 3D) Navier-Stokes equation demonstrate that the long-range entanglements from the MSR loss can be well modeled by the LordNet, yielding better accuracy and generalization ability than other neural networks. 
The results show that the Lordnet can be $40\times$ faster than traditional PDE solvers. 
In addition, LordNet outperforms other modern neural network architectures in accuracy and efficiency with the smallest parameter size.
\end{abstract}

\section{Introduction}
\label{introduction}
Partial differential equations (PDEs), imposing relations between the various partial derivatives of an unknown multivariable function, are ubiquitous in mathematically-oriented scientific fields, such as physics and engineering~\citep{landau2013course}. 
The ability to solve PDEs accurately and efficiently can empower us to understand the physical world deeply. 
More concretely, the goal of solving parametric PDE, i.e., a family of PDEs across a range of scenarios with different features, including domain geometries, coefficient functions, initial and boundary conditions, etc., is to find the solution operator that can map the parameter entities defining PDEs to corresponding solution functions. 
Conventional numerical solvers, such as finite difference methods (FDM) and finite element methods (FEM), are employed to form up solution operators based on space discretization. 
However, in many complex PDE systems requiring finer discretization, traditional solvers will be inevitably time-consuming, especially when solving boundary value problems or initial value problems with implicit methods.
 
More recently, deep learning-based methods have been successfully used to provide faster PDE solvers through approximating or enhancing conventional ones in many cases~\citep{guo2016convolutional,zhu2018bayesian,sirignano2018dgm,han2018solving,hsieh2018learning,raissi2019physics,bhatnagar2019prediction,bar2019learning,berner2020numerically,li2020fourier,li2020multipole,li2020neural,Um2020solver,pfaff2020learning,lu2021learning,wang2021learning,kochkov2021machine}. 
These neural-network-based (NN-based) PDE solvers can roughly be classified into three categories based on the role the neural network plays. 

Among all, one category of approaches takes the \textbf{Neural network as a component} to replace the component in conventional numerical solvers.
\citet{ling2016reynolds} learned representations of unclosed terms in Reynolds-averaged and large eddy simulation models of turbulence.
\citet{hsieh2018learning} learned a correction term in the iterative method of solving linear algebraic equations with convergence guarantees.
\citet{bar2019learning} learned coefficients of finite difference scheme from simulated data.
\citet{Um2020solver} learned a correction function of conventional PDE solvers to improve accuracy.
Similar to \cite{bar2019learning} and \cite{Um2020solver}, \citet{kochkov2021machine} learned a corrector and interpolation coefficients in finite volume methods for resolving sub-grid structure.
\citet{hermann2020deep} and \citet{pfau2020ab} proposed PauliNet and FermiNet, respectively, to approximate the wave-function for many-electron Schr\"odinger equation instead of conventional hand-crafted Ansatz in variational quantum Monte Carlo methods. 
\citet{list2022learned} proposed to model the turbulence with differentiable fluid solvers.

To further increase the efficiency of PDE solving, data-driven methods or pure NN-based solvers, which directly learn the solution operators, have attracted much attention since they could be orders of magnitude faster than traditional solvers. 
The approaches, taking the \textbf{Neural network as a solution}, use neural networks to approximate the solution of a specific PDE. 
\citet{sirignano2018dgm} and \citet{han2018solving} successfully approximated the solution of the nonlinear Black–Scholes equation for pricing financial derivatives and the Hamilton–Jacobi–Bellman equation in dynamic programming with neural networks.
Similarly, \citet{raissi2019physics} proposed physics-informed neural networks (PINN) to solve forward ~\citep{jin2021nsfnets} and inverse~\citep{raissi2020hidden} problems. 
They construct learning targets from PDEs but in a straightforward way. 
Thus, it could easily deal with irregular geometry and complex governing equations, and could infer with arbitrary resolution outputs. 
However, they directly substitute solution approximators into PDEs and take the residuals as loss functions, meanwhile treating definite conditions or data assimilating terms as regularization, leading to very sensitive weights to balance these terms. 
It should be further noted that the increase of the input dimension or the complexity of the expected represented function in large-scale problems will make the convergence of the neural networks hard and even prone to failure \citep{Moseley2021}. 
Due to the low-frequency bias of deep neural networks \citep{rahaman2019spectral}, PINNs also fail to solve PDEs with high-frequency or multi-scale structure \citep{cuomo2022scientific}. 
What’s more, as these approaches train a neural network to solve a specific PDE, they are usually much slower than conventional solvers except for very high-dimensional problems.

To overcome the aforementioned limitations, another category of approaches,  using the \textbf{Neural network as a solver}, learns a mapping from a parameter space to a solution space, so that it can solve a family of PDEs at a time, which is called parametric PDEs and also the problem we want to solve in this work.
Through offline training with pre-generated simulated data, most of them could infer faster than conventional solvers.
\citet{guo2016convolutional} and \citet{bhatnagar2019prediction} learned a surrogate model with CNN to predict the steady-state flow field from geometries presented by signed distance functions (SDF).
\citet{zhu2018bayesian} further used Bayesian CNN to improve the sample efficiency of learning surrogate models.
\citet{li2020multipole,li2020neural} proposed graph kernel networks to learn operators mapping from parameter spaces to solution spaces, which is invariant to different discretization.
\citet{li2020fourier} further proposed Fourier transformation as a graph kernel and obtained encouraging results in solving the Navier-Stokes equation.
In parallel, \citet{lu2021learning} proposed another approach called DeepONet to learn a mapping between the function spaces and successfully solved many parametric ODE/PDEs. 
Inspired by physics-informed neural networks (PINNs)~\citep{raissi2019physics},
\citet{wang2021learning} trained DeepONet in a physics-informed style and thus improved sample efficiency.
Additionally, \cite{pfaff2020learning} learned mesh-based simulations with graph networks and obtained impressive results in many applications. 
\citet{brandstetter2022message} proposed a neural solver based on neural message passing, representationally containing some classical methods, such as finite differences, finite volumes, and WENO schemes. 
\citet{brandstetter2022clifford} also proposed a Clifford neural network to consider the coupled relation between the various input physical fields to achieve better performance of the neural operator.

Particularly, as typical supervised learning tasks, these methods first collect data from numerical simulations and then use them for training neural networks mapping parameter entities to solution functions. 
Nevertheless, this supervised learning paradigm in fact imposes a chicken-egg dilemma in terms of efficiency. 
Specifically, while deep learning-based methods are originally employed to replace time-consuming numerical solvers, training a neural network model with high generalization ability, on the other hand, requires a large amount of data that has to be obtained through running time-consuming numerical solvers. 
This dilemma may substantially limit the usage of deep learning techniques in solving PDEs.

To avoid this dilemma, physics-constrained loss \citep{geneva_modeling_2020,Wandel2020,Wandel2021} can be used to train the neural networks. 
In general, the learning signal is obtained by evaluating the residual, i.e., to what extent the input parameter entities and output solutions satisfy the PDE. 
Such a learning signal, as a consequence, inspires us to learn the neural network solver directly from PDEs rather than relying on a large amount of simulated data laboriously collected by numerically solving parametric PDE. 
More precisely, we transform the PDE into algebraic equations with certain discretization schemes on collocation points and then define the physics-constrained loss with the residual error of the algebraic equations in the predicted solution at the collocation points, which we refer to as \textbf{M}ean \textbf{S}quare \textbf{R}esidual (MSR) loss. 

However, we find that most neural network architectures perform poorly for certain PDEs when trained with the MSR loss. 
We take a thorough rethinking of the internal mechanism of MSR loss and found that the long-range entanglements between collocation points, whose patterns vary in different PDEs, are the key to success.
But the inductive biases of most modern network architectures, such as the translation invariance, do not generalize well for different kinds of entanglements, yielding poor performance.
To this end, we design a more flexible neural network architecture to address this limitation, called LordNet (\textbf{Lo}w-\textbf{r}ank \textbf{d}ecomposition \textbf{Net}work). 
LordNet establishes the long-range entanglements by the low-rank approximation with simple matrix multiplications, which is tunable to model the entanglements of various PDEs with reduced computational cost.

As far as we know, we are the first to investigate the long-range entanglements inside the MSR loss and discuss the neural network design for it. 
In summary, our main contributions are three-fold:
\begin{itemize}
      \item We rethink the MSR loss and demonstrate the importance of long-range entanglements from MSR loss for neural PDE solver;
      \item We design a new and efficient network architecture called LordNet to model various patterns of the physical long-range entanglements introduced by the MSR loss;
      \item We evaluate the proposed method on two representative and crucial PDEs, Poisson's equation and (2D and 3D) Navier-Stokes equation, and achieve significant accuracy and efficiency improvements over several mainstream approaches.
\end{itemize}

The rest of this paper is organized as follows. 
We first introduce the problem and then describe our methods, including learning target and network architecture, in Section~\ref{methodology}.
To demonstrate the effectiveness and efficiency of the proposed method, Section~\ref{experiments} presents the experiment settings and results.
Finally, we discuss (Section~\ref{discussion}) and conclude in Section~\ref{conclusion}.

\section{Methodology} 
\label{methodology}
\subsection{Preliminaries}
Let $(\mathcal{A},\mathcal{U},\mathcal{V})$ be a triple of Banach spaces of function on a connected domain $\Omega\subseteq\mathbb{R}^n$ with boundary $\partial\Omega$, and $\mathcal{L}:\mathcal{A}\times\mathcal{U}\rightarrow\mathcal{V}$ be a linear or nonlinear differential operator, and $\mathcal{B}$ represents the boundary condition operator applied on $u$.
We consider the parametric PDE taking the form of
\begin{align}\label{eq:pdes}
    \mathcal{L}_a[u](x)&=0, \quad x\in\Omega \\
    \mathcal{B}[u](x)&=0, \quad x\in\partial\Omega
\end{align}
where $a\in\mathcal{A}$ denotes the parameters of the operator $\mathcal{L}$, such as coefficient functions, and $u\in\mathcal{U}$ is the corresponding unknown solution function.
We assume that, for any $a\in\mathcal{A}$, there exists a unique solution $u=\mathcal{F}(a)\in\mathcal{U}$ making Eq.~(\ref{eq:pdes}) satisfied, then $\mathcal{F}$ is the solution operator of Eq.~(\ref{eq:pdes}). 
Considering the Poisson's equation with the Dirichlet boundary condition, Eq.~(\ref{eq:pdes}) becomes
\begin{align}\label{eq:poisson-preli}
    \mathcal{L}_a[u](x) = \nabla^2 u(x) + a(x) &= 0, \quad x\in \Omega, \\
    \mathcal{B}[u](x) = u(x) &=0, \quad x\in \partial\Omega.
\end{align}

Since only few PDEs can be solved analytically, people have to resort to solving Eq.~(\ref{eq:pdes}) numerically.
Numerical solvers rely on converting PDEs into large-scale algebraic equations through discretization and then solving them iteratively.
For example, FDM discretizes $\Omega$ with $m_u$-point discretization $\{x_i\}^{m_u}_i\subset\Omega$, to convert the function $u$ into the vector form $\widehat{u} = \left[u(x_1),\cdots,u(x_{m_u})\right]$. 
Similarly, for the function $a$, we have the $m_{a}$-points discretization $\widehat{a}$. 
Note that the two point sets can be different, e.g., in the case of the staggered grid~\citep{harlow1965numerical}.
In this way, Eq.~(\ref{eq:pdes}) turns to $m_u$ algebra equations with respect to $\widehat{u}$ and $\widehat{a}$, formally,
\begin{equation}\label{eq:d_pdes}
    \widehat{\mathcal{L}}(\widehat{u},\widehat{a})=0,
\end{equation}
where $\widehat{\mathcal{L}}$ represents an algebraic operator determined by PDEs, definite conditions, and finite difference schemes.
When Eq.~(\ref{eq:pdes}) is a boundary value problem or an initial value problem integrated with implicit schemes, Eq.~(\ref{eq:d_pdes}) is usually a set of tightly coupled large-scale algebraic equations that are commonly solved by iterative methods, such as Krylov subspace methods~\citep{saad2003iterative} for the linear cases and Newton-Krylov methods~\citep{knoll2004jacobian} for the  nonlinear cases.
However, numerically solving $\widehat{u}$ from Eq.~(\ref{eq:d_pdes}) is not only slow and costly,
but also specific for a particular $a$, and thus we have to repeatedly solve such large-scale algebraic equations when $a$ changes.

To solve Eq.~(\ref{eq:pdes}) more efficiently, some previous work trained neural solvers following the conventional supervised learning paradigm.
Specifically, randomly sample the parameters $\{\widehat{a}\}_{i=1}^N$, and then generate the training dataset $\mathcal{D}=\{(\widehat{a}^i,\widehat{u}^i)\}_{i=1}^N$ by solving Eq.~(\ref{eq:d_pdes}) numerically for each $\widehat{a}^i$, and finally, train a neural network with a cost function such as the mean squared error (MSE) loss,
\begin{equation} \label{eqn:mse}
    l(\theta)=\frac{1}{N}\sum_{i=1}^N||f_{\theta}(\widehat{a}^i) - \widehat{u}^i||^2,
\end{equation}
where $f_{\theta}:\widehat{a}\mapsto\widehat{u}$ is the neural network parameterized by $\theta$.
In this way, given a specific $\widehat{a}$, the trained neural solver could infer the corresponding discrete solution function $\widehat{u}$ to make Eq.~(\ref{eq:d_pdes}) hold.

\label{prob_statement}
\subsection{Learning from PDEs} \label{loss}
It is easy to observe that the relation between network input $\widehat{a}$ and output $\widehat{u}$ has been completely presented by Eq.~(\ref{eq:d_pdes}).
Thus, it might be unnecessary to pre-generate simulated data by numerically solving Eq.~(\ref{eq:d_pdes}) and then minimize the MSE loss Eq.~(\ref{eqn:mse}) to train network $f_\theta$ making Eq.~(\ref{eq:d_pdes}) hold. 
Directly constructing loss functions from Eq.~(\ref{eq:d_pdes}) is more straightforward and can get rid of pre-generating simulated data. 
We define the mean square residual (MSR) of Eq.~(\ref{eq:d_pdes}) as the loss function, that is,
\begin{equation}
    l(\theta)=\frac{1}{N}\sum_{i=1}^N||\widehat{\mathcal{L}}(f_{\theta}(\widehat{a}^i), \widehat{a}^i)||^2.
\end{equation}
We can replace the mean square function with other cost functions, which goes beyond the discussion in this work, and we use the MSR loss throughout the paper.
In this way, we just need to evaluate Eq.~(\ref{eq:d_pdes}) instead of solving it, so that there is no simulated data anymore required, making the learning process jump out of the chicken-egg dilemma. 
Besides, the MSR loss often leads to a good generalization which we will show later.

In contrast, the supervised loss like MSE in Eq.~(\ref{eqn:mse}) explicitly supervises the model with the numerical results in the training dataset, which is essentially solved according to the physical constraints. With sufficiently massive training data, both approaches share the exact global optimum. However, it is often the case that the training data is limited because the numerical simulation is costly, and then the neural network fails to learn such physical information and overfits the training data. 

Next, we investigate what physical knowledge the MSR loss leads the model to learn.

\begin{figure*}[t]
  \centering
  \includegraphics[width=0.85\linewidth,trim= 0mm 0mm 0mm 0mm, clip]{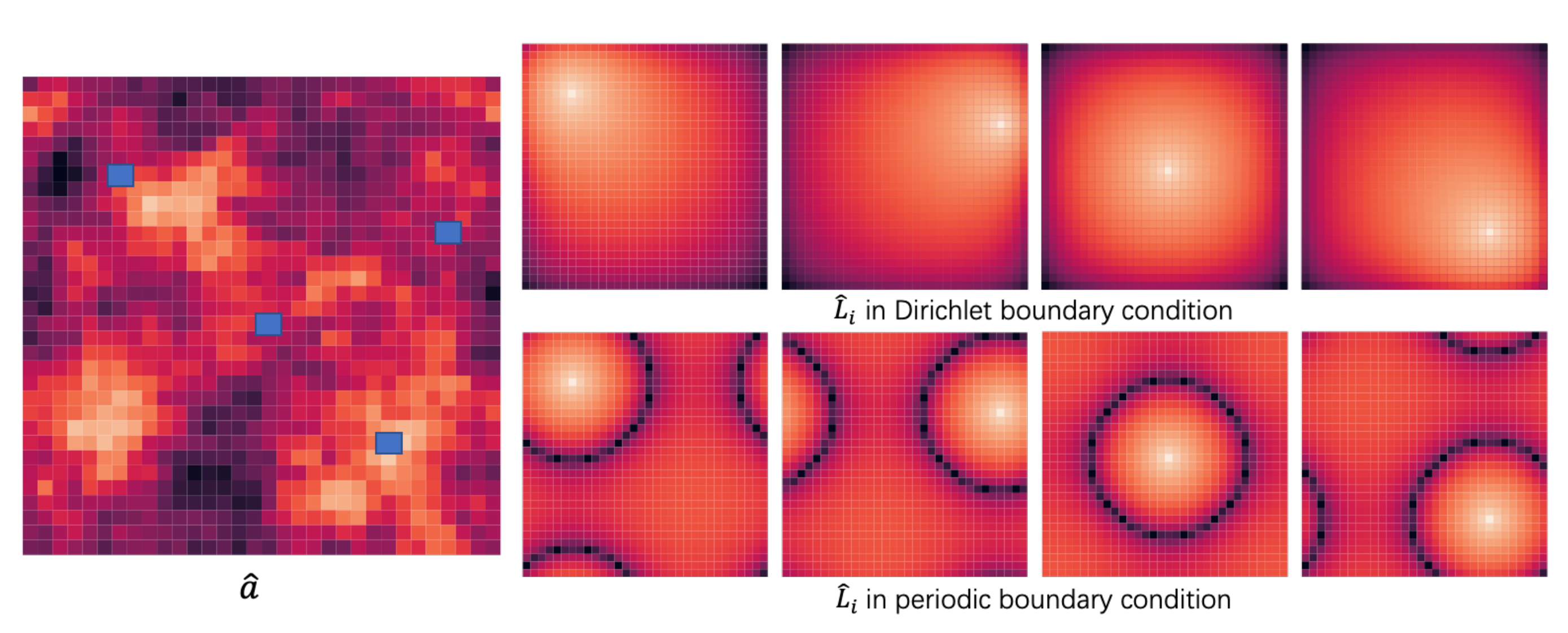}
  \caption{The diagram of Long-range entanglements: taking Poisson's equation as an example. The vectors $\hat{a}$ and $\hat{L}_i$ are reshaped into the matrix form to highlight the contribution of each point in the spatial domain to the blue dots.}
  \label{fig:poisson_entagle}
  \vspace{-6mm}
\end{figure*}

\textbf{Long-range entanglements.}
For boundary value problems or initial value problems integrated implicitly, Eq.~(\ref{eq:d_pdes}) is a set of tightly coupled algebraic equations; that is, each component of $\widehat{u}$ depends on many components of $\widehat{a}$.
Taking Poisson's equation~\ref{eq:poisson-preli} as an example of boundary value problems, suppose the domain $\Omega$ is square, and $u$ and $a$ are discretized with the grid of the shape $N\times N$. It is easy to get the linear equations in the matrix form $\widehat{\boldsymbol{L}}\widehat{u} = -\widehat{a}$. 
Denote the inverse of the matrix as $\widehat{\boldsymbol{L}}^{-1}$ and the $i$th row of the matrix as $\widehat{\boldsymbol{L}}^{-1}_i$. 
Then we have $-\widehat{\boldsymbol{L}}^{-1}_i \widehat{a} = \widehat{u}_i$, which shows how the sampled points of $a$ determine $u$ at a target point $x_i$.
The first line of Fig.~\ref{fig:poisson_entagle} shows $\log \left|\widehat{\boldsymbol{L}}^{-1}_i\right|$ for four choices of $i$ whose positions are marked in the heat map of $\widehat{a}$.
As you can see, even the grid points far away from the target points still have strong influences on the target points and cannot be ignored, the phenomenon of which is referred to as the {\it long-range entanglements} in this paper. 
The entanglements exist in many other PDEs. 
For example, the commonly used projection method for fluid dynamics simulation contains the procedure of solving Poisson's equation, which suffers a similar entanglement problem in certain boundary conditions. 

\subsection{LordNet} \label{network}
Although many modern neural network architectures have the capacity to model long-range entanglements, the de-facto architectures from other domains may not work well for solving PDEs, since the introduced inductive bias could be incompatible with various entanglements required by different PDEs. It is common that one neural network works well for a certain PDE, but works poorly for another, or even for the same one with a different boundary condition.
For example, CNNs~\citep{lecun1995convolutional} performs well in solving Poisson's equation with the periodic boundary condition, while performing poorly in that with the Dirichlet boundary condition. 
This is because under the periodic boundary condition, collocation points at different places are identical and the assumption of the translation invariance naturally holds, as you can see from the second line of Fig.~\ref{fig:poisson_entagle}. However, such symmetry is broken under the Dirichlet boundary condition. 
Additionally, graph neural networks(GNNs)~\citep{li2020multipole,li2020neural,pfaff2020learning} have been introduced for solving PDEs recently.
However, as \cite{li2020fourier} pointed out, GNNs fail to converge in some important cases, such as turbulence modeling. Because GNNs are originally designed for capturing local structures, it is hard to support long-range entanglements.
Transformer might be a possible choice since self-attention can resolve global entanglements and does not introduce too much inductive bias \citep{geneva2022transformers}.
However, the huge computational overhead of the global-wise attention for resolving global dependency will make neural networks less superior to conventional numerical methods.
In short, we need a lightweight network structure that can effectively resolve the long-range entanglements and introduces little inductive bias at the same time.

\begin{figure}[!htb]
    \centerline{\includegraphics[width=\textwidth]{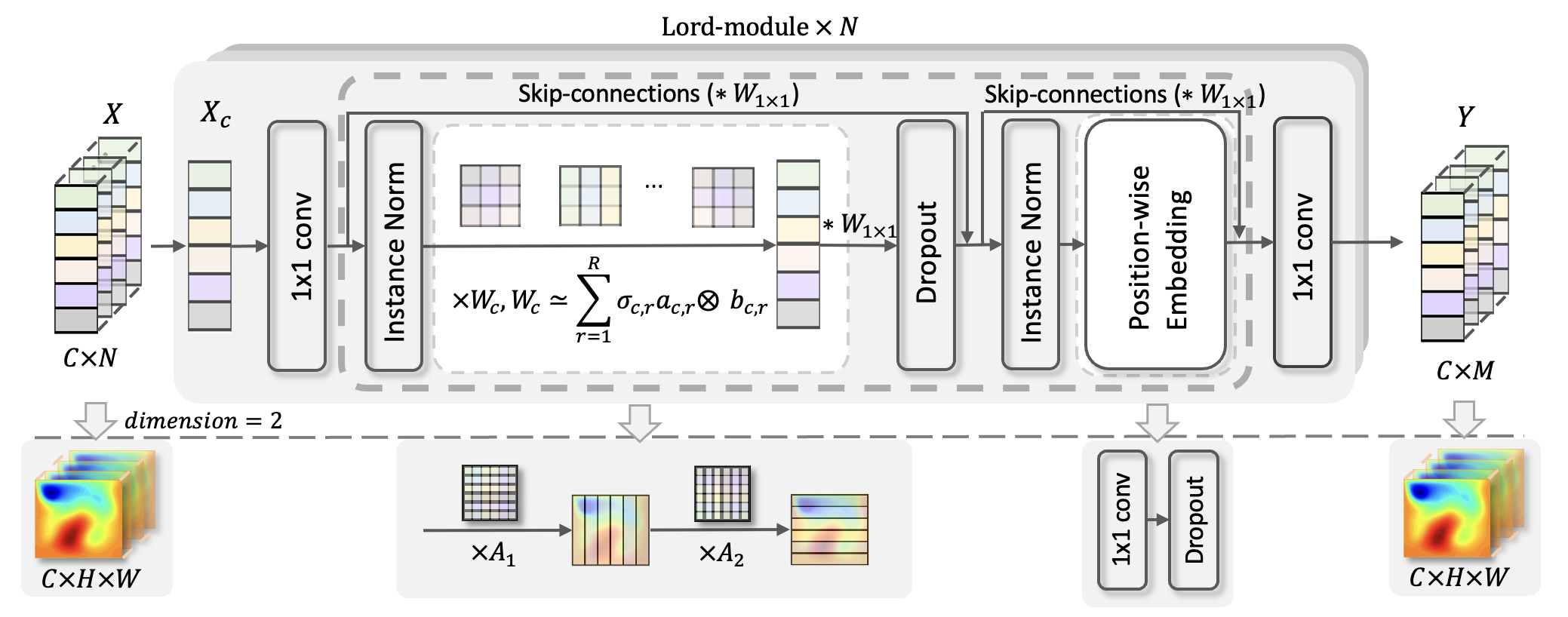}}
    \caption{An illustration of LordNet architecture.}
    \label{fig:lord}
\end{figure}

We design by first replacing the expensive multi-head self-attention in Transformer with a multi-channel fully-connected layer for two reasons. Firstly, the fully-connected layer is the most powerful but simple way to resolve linear long-range entanglements between input and output without any inductive bias. Secondly, it resembles the matrix operations in traditional solvers, i.e., matrix multiplication over the input, and we believe that such similarity gives the power to mimic the expensive numerical operations efficiently.
Specifically, a multi-channel fully-connected layer maps input $X\in\mathbb{R}^{C\times N}$ to output $Y\in\mathbb{R}^{C\times M}$ is defined as
\begin{equation}\label{eq:mcfc}
    Y_{c,m} = \sum_{n=1}^N W_{c,m,n} X_{c,n},
\end{equation}
where $W\in\mathbb{R}^{C\times M\times N}$ is the weight tensor, $c=1,2,\cdots,C$, $n=1,2,\cdots,N$ and $m=1,2,\cdots,M$ are the indices of the channel, input dimension, and output dimension, respectively.
Here, we linearly transform different channels independently, followed by a channel mixer without changing the dimension of the channel. 
It should be mentioned that for linear equations, e.g., Possion's equation, one layer is sufficient, and the optimal solution for $W$ is  $\widehat{\boldsymbol{L}}^{-1}_i$. 
However, the weight matrix $W$ is unacceptable in practice since $N$ and $M$ are usually large for discretizing $\Omega$.
Thus, we resort to a low-rank approximation to each channel of the weight matrix, that is,
\begin{equation}
    W_c \simeq \sum_{r=1}^R \sigma_{c,r} a_{c,r} \otimes b_{c,r},
\end{equation}
where $R$ is the rank of approximation, $\sigma_r$ are learnable singular values for $W_c$, $a_{c,r}\in\mathbb{R}^M$, and $b_{c,r}\in\mathbb{R}^N$ are corresponding left and right singular vectors, and $\otimes$ is the Kronecker product.
In this way, we could diminish the parameter size from $CMN$ to $CR(M+N)$.
We name the building block constructed this way as a Lord (\textbf{Lo}w-\textbf{r}ank \textbf{d}ecomposition ) module,
and a network consisting of Lord modules as LordNet.
In practice, the input $\widehat{a}$ with $N$-point discretization is first lifted to a higher dimensional representation $\mathbb{R}^{C\times N}$ by a point-wise transformation and then fed into the stacked Lord modules.
To better understand the network architecture, taking Lordnet for 2-dimensional (2D) PDE as an example, the whole architecture of LordNet is illustrated in Fig.~\ref{fig:lord}. Noted that $\ast W_{1\times1}$ means the 1$\times$1 convolution kernel and skip-connection here is different from the commonly used $(x+f(x))$, and we use $f_1(x)+f(x)$ as an alternative augmented shortcut, where $f_1$ is a 1$\times$1 convolution layer. We also compose the LordNet with the non-linear activation functions to model the non-linear operators.

When the discretization of $\Omega\subseteq\mathbb{R}^d$ is uniform,
the input matrix $X$ and output matrix $Y$ may be considered as high-dimensional tensors $\mathbb{R}^{C\times I_1 \times\cdots\times I_d}$ and $\mathbb{R}^{C\times O_1 \times\cdots\times O_d}$, respectively, where $I_1 \times\cdots\times I_d=N$ and $O_1 \times\cdots\times O_d=M$.
This structure can be leveraged to speed up the multi-channel fully-connected layer further.
Specifically, the rank-$R$ approximation of the weight tensor for each channel can be written as
\begin{equation} \label{eq:cp}
    W_c \simeq \sum_{r=1}^R \sigma_{c,r} a_{c,r,1} \otimes b_{c,r,1} \otimes \cdots \otimes a_{c,r,d} \otimes b_{c,r,d},
\end{equation}
where $a_{c,r,i}\in\mathbb{R}^{I_i}$ and $b_{c,r,i}\in\mathbb{R}^{O_i}$ for $i=1,\cdots,d$.
In this way, the parameter size can be further reduced to $CR\sum_{i=1}^d(I_i + O_i)$.
However, we empirically find that the approximation in Eq.~(\ref{eq:cp}) usually needs very large $R$ for good performance,
since the parameters are too few to resolve the relation between inputs and outputs.
Thus, we propose to approximate the weight tensor in a more representative way, that is,
\begin{equation} \label{eq:cp_m}
    W_c \simeq \sum_{r=1}^R \eta_{c,r} A_{c,r,1} \otimes A_{c,r,2} \otimes \dots \otimes A_{c,r,d},
\end{equation}
where $A_{c,r,i}\in\mathbb{R}^{I_i\times O_i}$ for $i=1,2,\cdots,d$.
Eq.~(\ref{eq:cp}) can be considered as a special case of Eq.~(\ref{eq:cp_m}) by letting $A_{c,r,i}$ be a rank-$1$ matrix with the decomposition $A_{c,r,i}=a_{c,r,i} \otimes b_{c,r,i}$,
thus the learning capability of Eq.~(\ref{eq:cp_m}) is much better than that of Eq.~(\ref{eq:cp}) so that we can retain much fewer components to approximate $W$. 
In practice, setting $R=1$ is often found to be sufficient for achieving satisfactory performance, and thus, it has been selected for use in the subsequent experiments conducted in this study.

Further, it should be noted that the decomposition in Eq.~(\ref{eq:cp_m}) is not only a mathematical approximation but also has a very clear physical meaning. We will discuss this with a 2D example, that is,
$X\in\mathbb{R}^{C \times I_1 \times I_2}$ and $Y\in\mathbb{R}^{C \times O_1 \times O_2}$.
In this case, the rank-$R$ approximation of the multi-channel linear transformation takes the form
\begin{equation}\label{eq:2d}
    Y_{c,o_1,o_2} = \sum_{r=1}^R \sum_{i_1=1}^{I_1} \sum_{i_2=1}^{I_2} A_{c,r,i_1,o_1} A_{c,r,i_2,o_2} X_{c,i_1,i_2},
\end{equation}
which indicates that, for each channel and rank, the input matrix left multiplies a weight matrix and then right multiplies another weight matrix.
The left multiplication first passes messages among positions row-wise,
and then the right multiplication passes messages among positions column-wise
so that all positions can efficiently communicate with each other in this way. Similarly, the implementation of 3-dimensional (3D) space is straightforward (Eq. \ref{eq:3d}). The main modification is adding a right multiplication with a different weight matrix, which passes messages among positions depth-wise.
\begin{equation}\label{eq:3d}
    Y_{c,o_1,o_2,o_3} = \sum_{r=1}^R \sum_{i_1=1}^{I_1} \sum_{i_2=1}^{I_2} \sum_{i_3=1}^{I_3} A_{c,r,i_1,o_1} A_{c,r,i_2,o_2} A_{c,r,i_3,o_3} X_{c,i_1,i_2,i_3},
\end{equation}

We notice that the 2D version of LordNet shares similarities with several works in computer vision. For example, \citep{tolstikhin2021mlp,lee2021fnet} find that simple MLP performs comparably with CNNs and Transformer in computer vision and natural language processing tasks, respectively.
The architectures they used can be seen as specific instances of LordNet,
so we believe that, besides solving PDEs, LordNet might also perform well in traditional deep learning applications.
We leave it for our future work.

Besides low-rank approximation to the fully-connected layer, Fourier transform is also an effective and efficient way to model long-range entanglements. 
For example, we empirically find that Fourier neural operator, which models the entanglements with Fast Fourier Transform (FFT), also performs well in multiple cases. 
However, FFT originally forms samples of a periodic spectrum, and for problems with non-periodic boundary conditions, FNO has to add zero-paddings to correct the inductive bias. 
In addition, it ignores the high-frequency parts, which may hurt the performance of the neural solvers in some PDEs. 
As will be shown in our experiments, LordNet performs better than it in most of the cases. 

\section{Experiments} \label{experiments}
In this section, we will showcase LordNet's capability to model various long-range entanglements. 
We first highlight the importance of modeling for long-range entanglements. Then we demonstrate that LordNet gives the flexibility to model various entanglements in 2D/3D fluid problems both efficiently and accurately and outperforms other neural PDE solvers in both the physics-constrained manner and the supervised manner.  
We choose two representative and crucial PDEs, Poisson's equation and Navier-Stokes equation, as the testbeds.
\subsection{2D Possion's Equation}
As aforementioned, long-range entanglements exist in the MSR loss. In this experiment, we will investigate how the inductive bias of a neural network influences the performance in PDEs with different physical entanglements and the flexibility of LordNet to deal with various long-range entanglements. 
Following the discussion in Section~\ref{network}, we use time-independent Poisson's equation to test the periodic and Dirichlet boundary with both CNN and LordNet. 
The formula with the Dirichlet boundary condition is described in Eq.~(\ref{eq:poisson-preli}), which is used in our experiments. 
For the periodic boundary condition, the domain is cyclic at both dimensions. 
To model the long-range entanglements, we construct the CNN by stacking multiple dilated convolutional layers to increase the receptive field to the whole domain.
As shown in Table~\ref{tab:entanglemnt_exp}, both CNN and LordNet achieve a low relative error for the periodic boundary condition. 
However, CNN works poorly for the Dirichlet boundary condition where the entanglements are not translation invariant. Because LordNet is flexible to model various long-range entanglements, it performs well in both cases.
\begin{table}[!htb]
    \centering
    \caption{Comparison on Poisson's equation in terms of relative error and solution time (ms).}
    \setlength{\tabcolsep}{0.02\columnwidth}
    \begin{tabular}{c|c|cc|cc}
         \toprule
        \multirow{2}{*}{Boundary} & \multirow{2}{*}{Network} & \multicolumn{2}{c|}{$n=32$} & \multicolumn{2}{c}{$n=128$} \\ 
        & & Error & Time &  Error & Time \\
        \midrule
        Periodic & Conv & 0.00057$\pm$0.00005 & 0.8 &0.00483$\pm$0.00045 & 1.3\\
         Periodic & LordNet & 0.00051$\pm$0.00006 & 0.5 & 0.00379$\pm$0.00031& 1.2\\
         \midrule
         Dirichlet & Conv & 0.73042$\pm$0.00707& 0.8 & 1.01302$\pm$0.02233 & 1.3 \\
         Dirichlet & LordNet & 0.00265$\pm$0.00015 & 0.5 & 0.06937$\pm$0.00914 & 1.2 \\
        \bottomrule
    \end{tabular}
    \label{tab:entanglemnt_exp}
\end{table}

\subsection{2D Navier-Stokes Equation.}
Now that we see the importance of modeling the long-range entanglements and the capability and flexibility of LordNet for various long-range entanglements in the time-independent Poisson's equation, we will further demonstrate the superiority of LordNet compared to other neural PDE solvers in the more challenging time-dependent and non-linear Navier-Stokes equation. Besides convolutional neural networks like ResNet, we test the other two, FNO and transformer, which are supposed to be able to model various long-range entanglements.
We also test them in supervised training with plenty of data for a more comprehensive comparison.

\subsubsection{Task Descriptions and Evaluation Metrics}
We consider the 2-dimensional Navier-Stokes equation in vorticity-stream function form as follows,
\begin{align}\label{eq:nse}
    \frac{\partial\omega}{\partial t} = -\frac{\partial \psi}{\partial y}\frac{\partial \omega}{\partial x} + \frac{\partial \psi}{\partial x}\frac{\partial \omega}{\partial y} &+ \frac{1}{\text{Re}}\left(\frac{\partial^2 \omega}{\partial x^2} + \frac{\partial^2 \omega}{\partial y^2}\right) \\ \label{eq:nse1}
        \frac{\partial^2 \psi}{\partial x^2} + \frac{\partial^2 \psi}{\partial y^2} &= - \omega  
\end{align}
\label{eq:nse2}
where $\omega:[0, T]\times (0, 1)^2 \rightarrow \mathbb{R}$ is the vorticity function, $\psi:[0, T]\times (0, 1)^2 \rightarrow \mathbb{R}$ is the stream function, $\text{Re} \in \mathbb{R}_+$ is the Reynolds number. 
As a typical case of the time-dependent PDE, the neural network is used to predict the stream function in discrete timesteps in an auto-regressive manner. That is, the input of the neural network is the stream function at a certain timestep, and the output is that of the next timestep.
We consider the periodic boundary condition and Lid-driven cavity boundary condition.
The periodic boundary condition sets the domain cyclic in both dimensions. In this case, $T$ is 2, and the timestep is $0.01$. 
The Lid-driven cavity boundary condition~\citep{Zienki2006} is another well-known benchmark for viscous incompressible fluid flow, which is considered more complicated to solve than the periodic one. Specifically, the fluid acts in a cavity consisting of three rigid walls with no-slip conditions and a lid moving with a steady tangential velocity, which is set to 1 in our experiments. In this case, we evaluate for a longer time, i.e., $T$ is 27 with the timestep $1\mathrm{e}-2$.
In both conditions, we set the viscosity $\nu=1\mathrm{e}-3$ and fix the resolution to 64$\times$64. 
\subsubsection{Model Settings and Computational Runtime}
\textbf{Neural network baselines.} 
We choose the other three representative neural network architectures in our experiments. 
\textit{Resnet}: Resnet~\citep{he2016deep} is famous and widely used in computer vision tasks. We use the residual blocks implemented in PyTorch and change the batch normalization to layer normalization for better performance. 
\textit{Swin Transformer}: Swin Transformer~\citep{liu2021swin} leverages the Transformer architecture to handle computer vision tasks. 
It is popular and outperforms existing neural networks with a large gap in computer vision tasks. 
We directly use the implementation in its open-sourced project. 
\textit{FNO}: Fourier neural operator is proposed in \citep{li2020fourier} that also solves parametric PDEs.
\textit{LordNet}: LordNet with two Lord modules as illustrated in Fig.\ref{fig:lord}.

\textbf{Numerical solvers and other configurations.} We implement GPU-accelerated FDM solvers with the central differencing scheme for all cases as the numerical baselines, where the conjugate gradient method is used to solve the sparse linear algebraic equations. 
These numerical solvers are also used to collect training data and ground truth for tests.
In all experiments, we construct MSR loss functions with the same finite difference schemes as numerical solvers.
Additionally, we trained all neural networks on a V100 GPU with Adam optimizer and decayed learning rates.
The relative error is the evaluation metric for all experiments as the accuracy measurement, which is the Euclidean distance from the prediction to the ground truth divided by the Euclidean norm of the ground truth. For the Navier-Stokes equation, we calculate two types of relative error, the one denoted as `Error-1' is the one-step relative error, and the other denoted as `Error-mul' is the accumulated error of the terminal state when the model auto-regressively predicts along the multiple timesteps.
Please refer to \ref{app:numerical} and~\ref{app:exp_detail} for more details about the experiments.

\subsubsection{Performance Comparison}
\textbf{Lid-driven cavity.} We conduct a comprehensive study on the performance of different neural networks using the Navier-Stokes equation with the Lid-driven cavity boundary condition. 
We test with three other representative neural network architectures, i.e., ResNet, Swin Transformer, and FNO. 
The simulation time is 27, which is 2700 timesteps in total. 
The quantitative errors include Error-1 and Error-2700, describing the accuracy of the simulation.
Especially the accumulation effect in the sequential prediction as long as 2700 timesteps requires the neural network to be very accurate at every timestep of the sequence.
To prevent the generalization problem of MSE loss due to the out-of-distribution but test the generalization of MSR loss, the training data of the MSE loss cover the whole trajectories of 2700 timesteps, while for the MSR loss, we still only leverage the data of the initial conditions.

As you can see from Table~\ref{tab:loss}, the single-timestep error of both FNO and LordNet is below 1e-4, and the accumulated relative error of both FNO and LordNet trained with the MSR loss is controlled successfully under 5 percent, while all the other experiments fail even if the single-timestep relative error already reaches 0.1 percent or lower.
But LordNet is still slightly better than FNO, because FNO ignores the high-frequency parts (high-frequency truncation), which we believe is the main reason why FNO is not as good as LordNet in this case.
As we introduced previously, we believe that transformer, Fourier transform, and low-rank approximation are effective ways to model long-range entanglements. 
However, we find that Swin Transformer performs poorly in this setting.
One possible reason is that the patch-wise self-attention is too coarse to model the ‘pixel-wise’ entanglements. 
The architecture based on the attention mechanism should be re-designed for the PDE-solving tasks.
\begin{table}[t]
    \centering
    \caption{Comparison on Navier-Stokes equation with Lid-driven cavity boundary condition.}
    \setlength{\tabcolsep}{0.005\columnwidth}
    \begin{tabular}{c|c|cccc}
        \toprule
         \multirow{2}{*}{Method} & \multirow{2}{*}{Network} & \multirow{2}{*}{Error-1} & \multirow{2}{*}{Error-2700} & Inference time & Parameter  \\
         & & & &(ms) &size\\
         \midrule
         FDM & -- & -- & -- & 56.731 & -- \\
         \midrule
         \multirow{4}{*}{MSE}& ResNet & 0.000879$\pm$0.000071 & 0.8621$\pm$0.0109 & 5.131 & 1.21M \\
         & Swin Transformer & 0.000546$\pm$0.000063 & 0.6877$\pm$0.0148 & 23.011 & 59.82M \\
         & FNO & 0.000900$\pm$0.000013 & 0.6831$\pm$0.0161 & 1.659 & 9.46M \\ 
         & LordNet & 0.000738$\pm$0.000049 & 0.5172$\pm$0.0138 & 1.409 & 1.15M \\
         \midrule
         \multirow{ 4}{*}{MSR}& ResNet & 0.000432$\pm$0.000066 & 0.4004$\pm$0.0175 & 5.131 & 1.21M \\
         & Swin Transformer & 0.000529$\pm$0.000043 & 0.4410$\pm$0.0206 & 23.011 & 59.82M \\
         & FNO & 0.000089$\pm$0.000005 & 0.0429$\pm$0.0034 & 1.659 & 9.46M \\
         & LordNet & 0.000071$\pm$0.000009 & 0.0284$\pm$0.0036 & 1.409 & 1.15M\\
        \bottomrule
    \end{tabular}
    \label{tab:loss}
\end{table}

As for the comparison between various loss functions, there is an obvious performance gap between the neural networks trained with the MSE loss and those with the MSR loss, which indicates the effectiveness of the physical information in the MSR loss. But even with the MSE loss function, we still find that with insufficient data, LordNet  outperforms other neural networks.

\begin{figure*}[!htb]
  \centering
  \includegraphics[width=0.85\linewidth,trim= 0mm 0mm 0mm 0mm, clip]{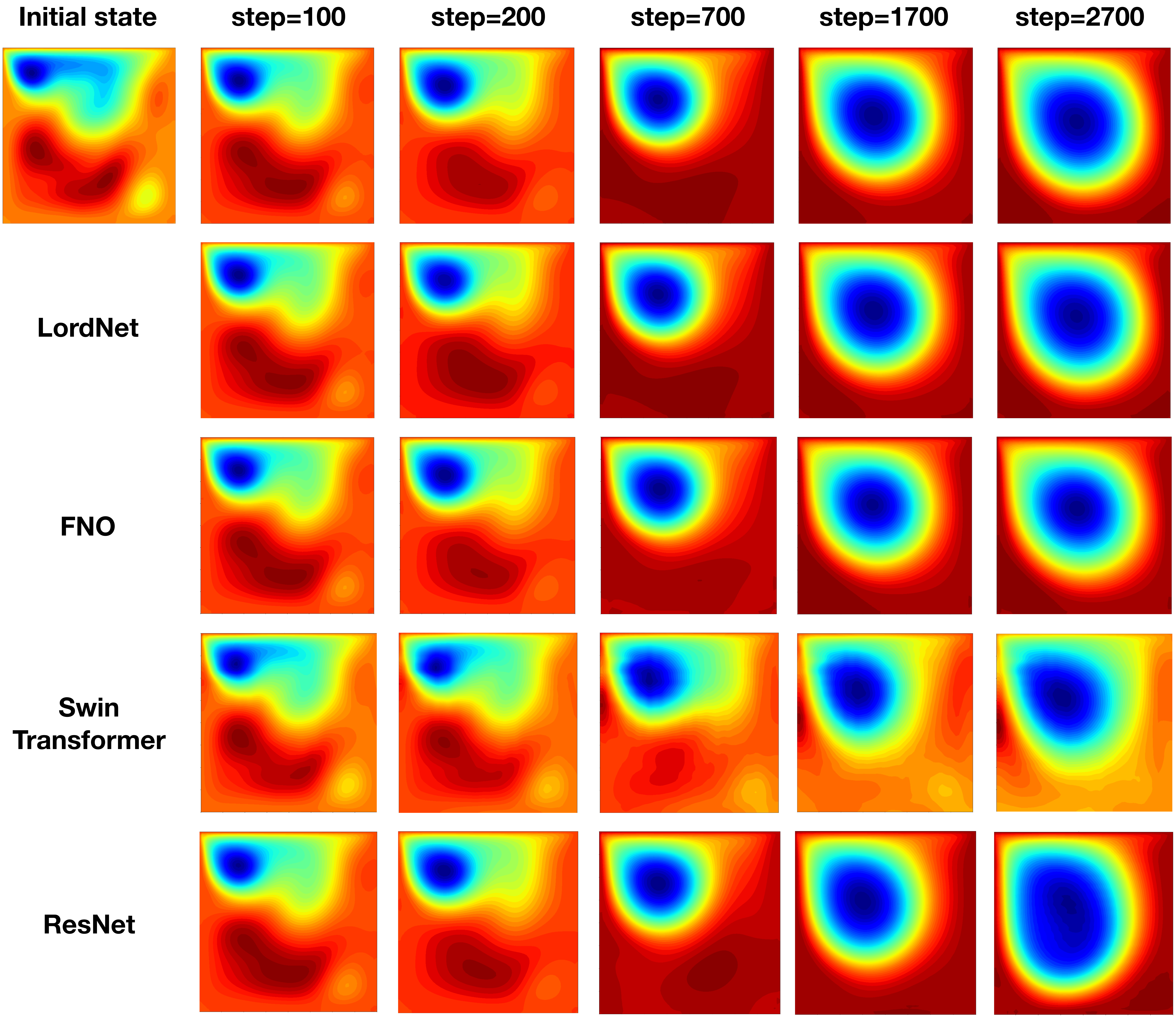}
  \caption{We present solutions to the Navier-Stokes equations in a lid-driven cavity context, employing MSR loss across various neural network architectures. The top row displays the numerical simulation results, while the subsequent rows from top to bottom illustrate the predictions generated by LordNet, FNO, Swin Transformer, and ResNet, respectively.}
  \label{fig:nse}
\end{figure*}

Besides, among all these neural networks, LordNet achieves the best result with the fewest parameters and the fastest inference speed. 
As Figure~\ref{fig:nse} shows, the simulation result by LordNet is nearly the same as the FDM solution and stabilizes to the same steady state as time goes on.
Compared to the GPU-accelerated FDM solution, LordNet has over 40 times speed-up. 

\textbf{Periodic boundary condition.} To further compare the FNO and LordNet, we leverage the Navier-Stokes equation with the periodic boundary condition to test.
For the MSE loss, we collect the training data by first sampling multiple initial conditions from a random distribution and then solving these PDEs with the FDM solver.
We test the relative error on unseen initial conditions sampled from the same distribution within the time length of 2.
To evaluate the generalization ability in this case, we add the experiment where the simulation trajectories for training only cover the period from 0 to 0.25, and the neural solvers are optimized with MSE loss.
To be fair, the size of the training set is 6000 for both experiments. From the result in Table~\ref{tab:nse}, we find with MSR loss, the performances of FNO and LordNet are almost the same because FNO is well-designed and can handle the periodic boundary conditions perfectly via FFT. But we observe that the LordNet using MSE loss with the data from 0 to 0.25 outperforms the FNO even with the data from 0 to 2.0. This shows the data volumes requirement for LordNet training is lower compared to FNO. One reason is that having fewer parameters in a model can make it more generalizable to new data because it is less likely to memorize the training data.
\begin{table}
    \centering
    \caption{Comparison on Navier-Stokes equation with periodic boundary condition.}
    \setlength{\tabcolsep}{0.02\columnwidth}
    \begin{tabular}{c|c|c|cc}
         \toprule
         \multirow{2}{*}{Method} & 
         \multirow{2}{*}{Network} &
         \multirow{2}{*}{Parameters}  &
         \multicolumn{2}{c}{$\nu=1 \mathrm{e}-3,T=2$} \\
         &&& Error-1 & Error-200. \\
         \midrule
         MSE (0\textasciitilde0.25) & LordNet & 1.15 M & 0.0000177$\pm$0.0000021 &0.00286$\pm$0.00023 \\
         MSE (0\textasciitilde0.25) & FNO & 9.46 M &  0.0000314$\pm$0.0000017  & 0.00391$\pm$0.00006  \\
         \midrule
         MSE (0\textasciitilde2) & LordNet & 1.15 M & 0.0000269$\pm$0.0000003 &0.00202$\pm$ 0.00008 \\
         MSE (0\textasciitilde2) & FNO & 9.46 M & 0.0000307$\pm$0.0000005  & 0.00234$\pm$0.00017 \\
         \midrule
         MSR (Initial only) & LordNet & 1.15 M & 0.0000092$\pm$0.0000006 &0.00247$\pm$0.00015\\
         MSR (Initial only) & FNO & 9.46 M & 0.0000099$\pm$0.0000010 & 0.00225$\pm$0.00012\\
        \bottomrule
    \end{tabular}
    \label{tab:nse}
\end{table}

\subsection{LordNet on a 3-dimensional fluid problem}
The proposed LordNet has shown priority in both accuracy and efficiency in the above experiments. 
But for real-world problems, 2D simulations are not sufficient to describe the general 3D fluid behavior learning \citep{lienen2023generative}. Thus, learning fluid dynamics in 3D is an evitable task and is always a challenge due to the increasing computational complexity and memory cost, as well as degrees of freedom for 3D fluid motion. 
Then in this section, we further demonstrate the priority of the LordNet in 3D fluid problems, where we use the LordNet in 3D versions. For the 3D experiments, we trained the neural networks on an A100 GPU with an Adam optimizer. 

The case used here is also built on the Navier-Stokes equation but in the 3D domain.
Specifically, we choose the same case described in \citep{wandel_teaching_2021} to test, in which they train a neural PDE solver from scratch (without training data) to simulate various fluid phenomena such as the Magnus effect or Kármán vortex streets. 
For the later fair comparison with benchmarks, we consider the same form of 3D Navier-Stokes equation as follows,
\begin{align}
\label{eq:nse-3d}
\rho\left(\frac{\partial \vec{v}}{\partial t}+(\vec{v} \cdot \nabla) \vec{v}\right) &=-\nabla p+\mu \Delta \vec{v}+\vec{f}\\
\nabla \cdot \vec{v} &=0
\end{align}
where $\vec{v}$ is the fluid velocity field, $p$ is the pressure field, $\mu$ is the viscosity, and $\vec{f}$ is the external force. In this case, the external force is set to 0, and the velocity satisfies the Dirichlet boundary conditions at the boundary of the domain, where $\vec{v}=\vec{v}_d$. 

\begin{figure}
    \centering
    \includegraphics[width=0.85\textwidth]{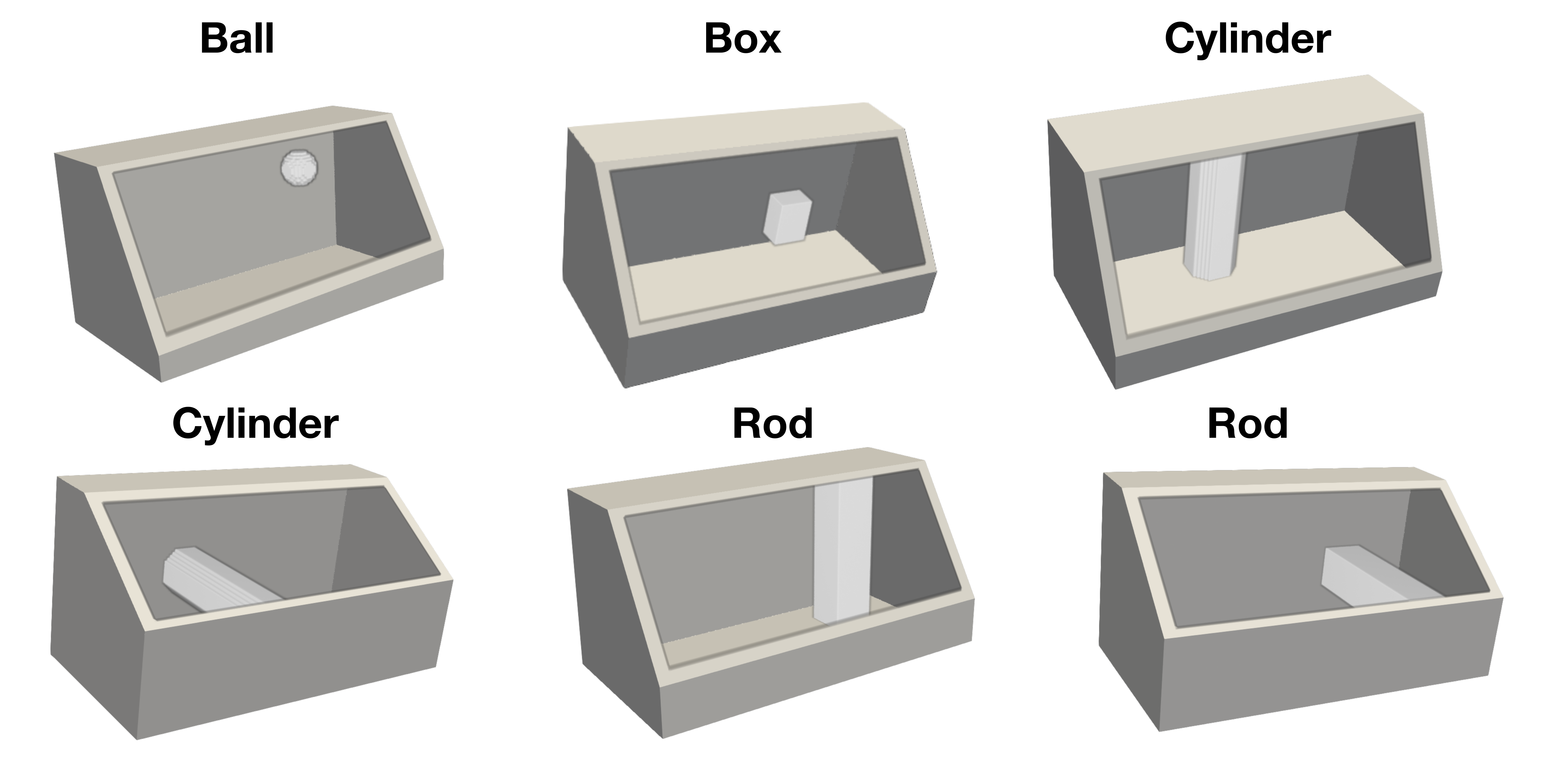}
    \caption{Randomly selected examples of training domains consisting of 128$\times$64$\times$64 voxels. The locations of the obstacles are also randomly chosen, and the inflow and outflow boundaries are on the left and right sides of the domains.}
    \label{fig:training}
\end{figure}

We also use the neural PDE solver to learn fluid dynamics in an auto-regressive manner. 
As the learned solutions are hard to guarantee the projections onto divergence-free parts, resulting in a violation of incompressibility within the domain \citep{tompson2017accelerating}, we follow the way of \citep{wandel_teaching_2021} to ensure incompressibility.
We set a vector potential $\vec{a}$ and let $\vec{v}=\nabla\times\vec{a}$, and then use the neural PDE solver to map the fluid states including $\vec{a}$ and $p$ at time point $t$ to them at time point $t+\Delta t$. 
Here the training pipeline is slightly different from the previous settings with the initial states only for the physics-constrained learning or the data-label pairs for the supervised learning, we train the network from scratch using the MSR of equation~\ref{eq:nse-3d}. 
We initialize a data pool including vector potential $\vec{a}$, pressure $p$, random shapes of interior domains, random boundary conditions, random viscosity $\mu$, and random fluid density $\rho$, in which we set the initial $\vec{a}$ and $p$ to 0. The number of voxels in the domain is 128$\times$64$\times$64, and there exist different shapes of obstacles inside the domain (Figure~\ref{fig:training}). 
During the training, we constantly update the training samples in the pool by means of the predictions with the neural networks and also reinitialize the pool regularly. 
As the training goes on, the training samples include states belonging to different trajectories and different time points, which are beneficial for the training to achieve a better generalization. 
More details can be seen in \ref{app:3d}.

\begin{figure}
    \centering
    \includegraphics[width=0.85\textwidth]{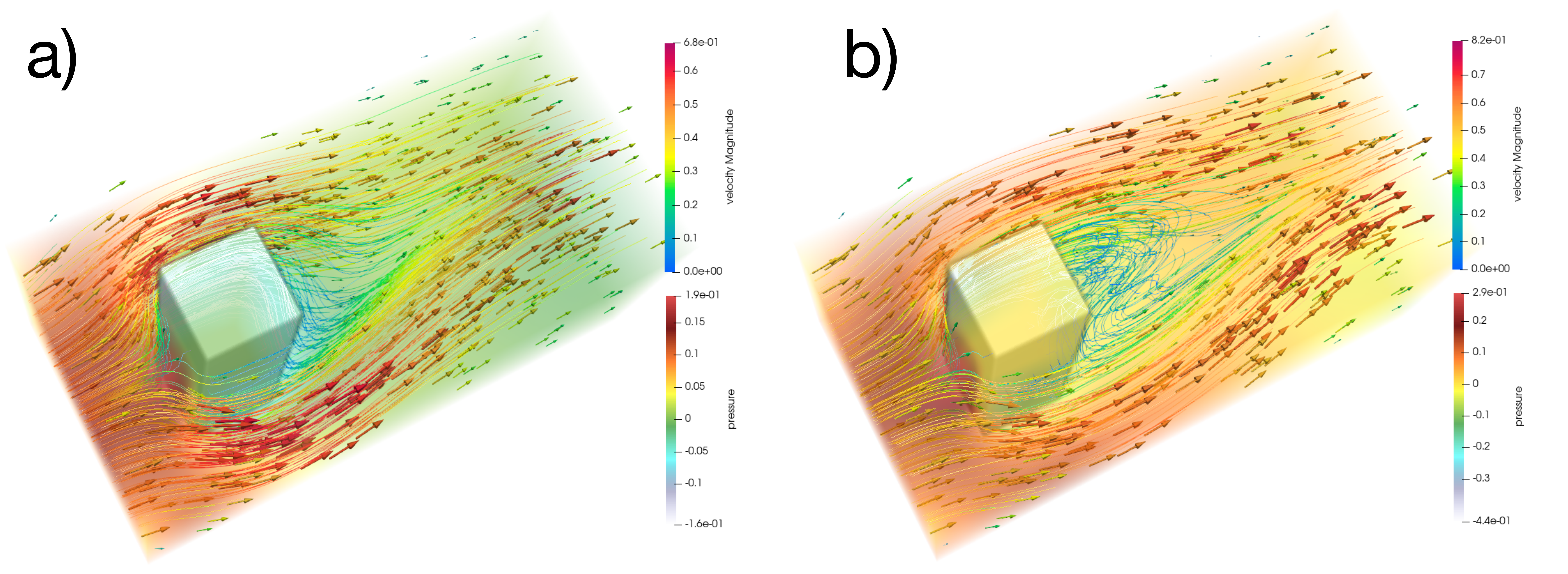}
    \caption{Streamlines and pressure fields of flow around a square rod at different Reynolds numbers. The diameter of the rod is 16. a) is the result of $\rho=0.5, \mu=3.0$, resulting in a Reynolds number of 48, while b) is the result of $rho=0.1, \mu=8.0$, resulting in a Reynolds number of 640. The results are generated by LordNet.}
    \label{fig:lordnet_eval}
\end{figure}

\begin{figure}
    \centering
    \includegraphics[width=0.85\textwidth]{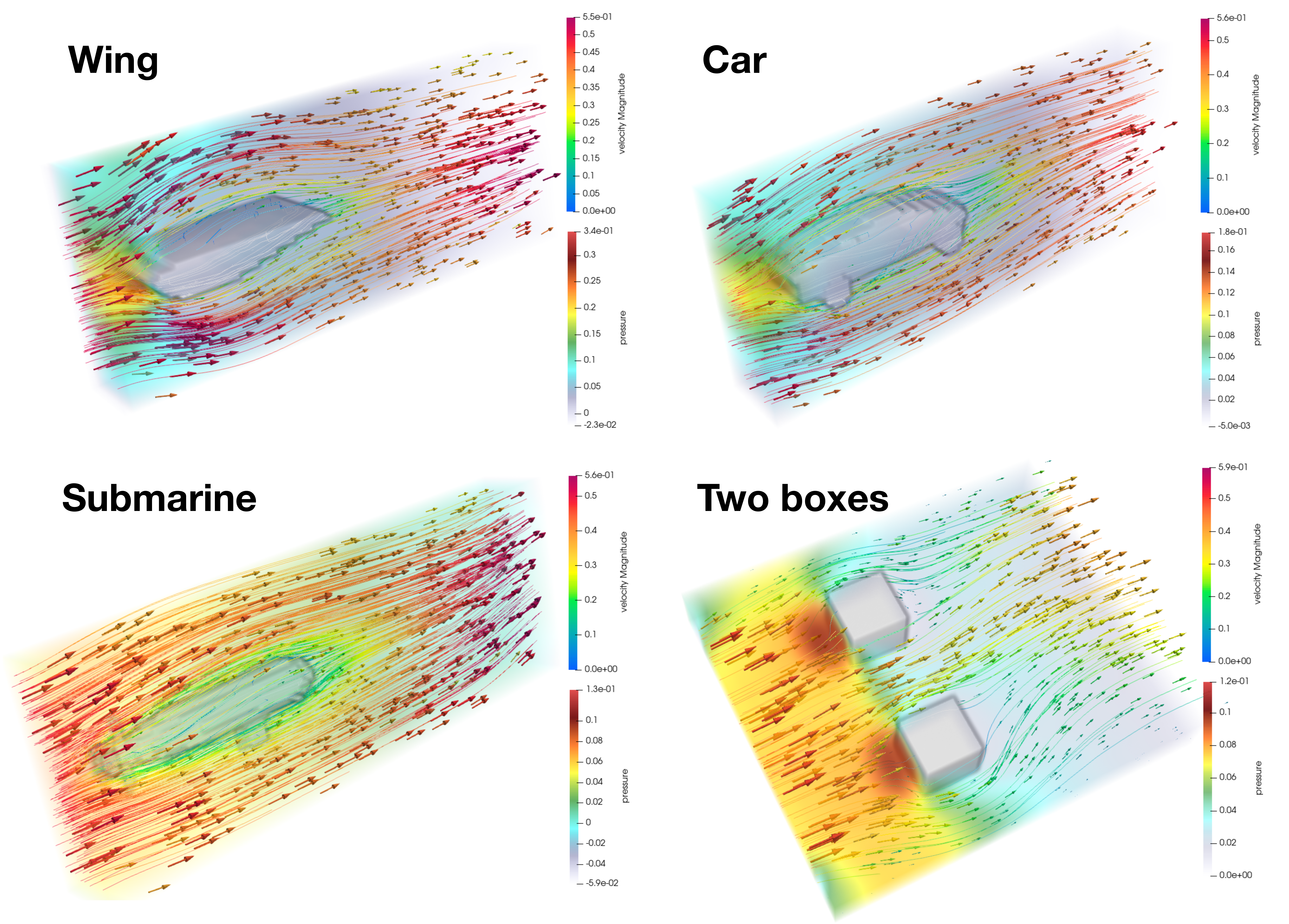}
    \caption{Generalization examples for objects whose shapes are not seen during training including wing, car, submarine, and two boxes, where $rho=0.5, \mu=2.0$. The results are generated by LordNet.}
    \label{fig:lordnet_general}
\end{figure}

As shown in Figure~\ref{fig:lordnet_eval}, the trained LordNet generates reasonable simulations in cases of laminar flow and turbulent flow. 
We further test its generalization on different boundaries, which are never seen during the training. Figure~\ref{fig:lordnet_general} demonstrates that with training on limited simple boundaries, the LordNet can generalize well on complex unseen boundaries.

\begin{figure}
    \centering
    \includegraphics[width=1.0\textwidth]{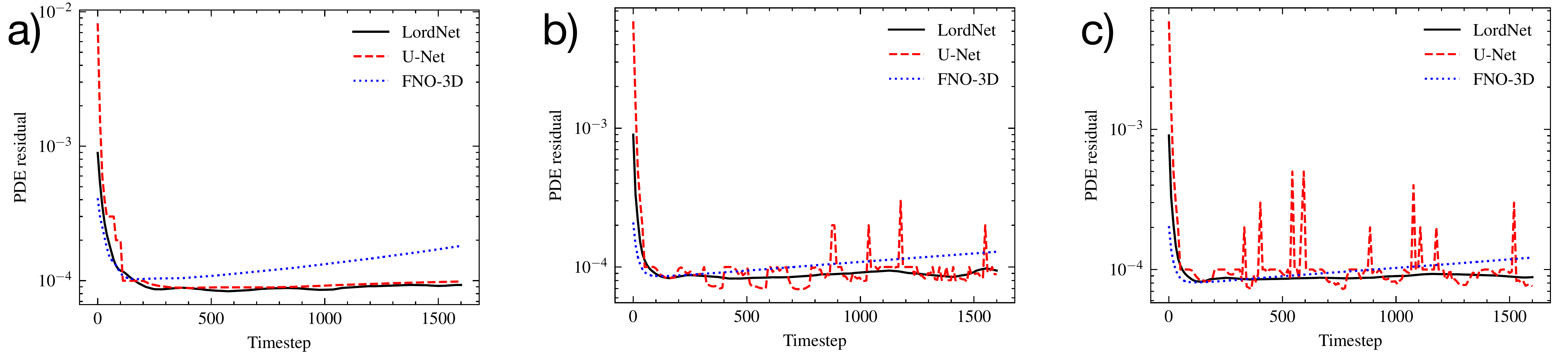}
    \caption{Quantitative comparison for stability and performance of different neural networks applied to the in-distribution (a) and the out-of-distribution (b and c) boundary shapes. a) is the test for one box obstacle, b) and c) are the tests for the wing and car shapes, respectively.}
    \label{fig:wing_and_car}
\end{figure}

To quantify the performance of the LordNet, and also demonstrate the superiority of the LordNet in this case compared to the benchmark Unet\citep{wandel_teaching_2021} and FNO\citep{li2020fourier}, we use the same benchmark shown in \citep{wandel_teaching_2021}. This is the case with a single box obstacle belonging to the in-distribution test.
To compare their generalization ability, we still further test these neural networks in the cases with out-of-distribution obstacles, e.g., wing-shape and car-shape obstacles (Figure~\ref{fig:lordnet_general}).
Noted that here, we train the neural networks from scratch, where we do not have any simulated data for training as well as testing.
For a fair comparison, we use the same ways of quantifying as shown in \citet{wandel_teaching_2021}, the accuracy with respect to PDE residual, to evaluate the model.
Figure~\ref{fig:wing_and_car} are the comparisons of the evaluation results along the timestep with different models, in which the smaller PDE residual means that the neural PDE solver is closer to the numerical finite-different method, and that is to say, it is more accurate. 
We find that they all perform well when dealing with in-distribution boundary shapes, and LordNet shows a slight edge in effectiveness. However, Unet has some fluctuations in out-of-distribution boundary shapes, while LordNet is still the most stable and accurate. In all cases, the error accumulation of FNO is apparent compared to others. 

Furthermore, as shown in Table~\ref{tab:flow_ap_3d}, the LordNet achieves the lowest PDE residual with the fewest parameters. 
Meanwhile, the inference time of the LordNet on an A100 GPU card is less than other networks.
\begin{table}[!htb]
    \centering
    \caption{Comparison on 3D Navier-Stokes equation.}
    \setlength{\tabcolsep}{0.025\columnwidth}
    \begin{tabular}{c|c|c}
        \toprule
         Network & Inference time & Parameter size \\
         \midrule
         Unet &  5.34 ms & 29.03 M  \\
         FNO-3D & 3.32 ms & 226.52 M  \\
         LordNet &  2.84 ms & 3.26 M \\
        \bottomrule
    \end{tabular}
    \label{tab:flow_ap_3d}
\end{table}

\section{Discussion}\label{discussion}
In this paper, we investigate the long-range entanglements of MSR loss and the network design for it. Besides the MSR loss, the physics-informed loss, like DGM~\citep{sirignano2018dgm} and PINN~\citep{raissi2019physics},
trained a network to approximate the solution function $u$ mapping from $\Omega$ to $\mathbb{R}$ given a specific $a$ rather than an operator mapping from $\mathcal{A}$ to $\mathcal{U}$.
They can be seen as another way to learn from PDEs, where the derivatives of the function to be sought in the PDE are analytically replaced by the derivatives of the neural network. 
Although the physics-informed loss can also generalize better compared to supervised training, its internal mechanism is not obvious (for PINNs whose derivates are obtained via automatic differentiation and built on MLP). The long-range entanglements and the principle of network design are not suitable for this framework.

Meanwhile, the LordNet deals with uniform grid points and its adaption to complex geometry needs further development, and we leave this to the future.

\section{Conclusion} \label{conclusion}
In this paper, we empirically demonstrate that the long-range entanglements from PDE loss (MSR) are the key to the performance of physics-constrained learning.
To model the long-range entanglements, we further propose a \textbf{Lo}w-\textbf{r}ank \textbf{d}ecomposition \textbf{Net}work (LordNet), yielding a faster and more accurate neural PDE solver.
The experiments on solving Poisson’s equation and Navier-Stokes equation demonstrate that the proposed LordNet efficiently models the long-range entanglements and outperforms modern neural network architectures with fewer parameters. 
We also found that even in a supervised manner, the LordNet still has a better generalization and accuracy in PDE solving with limited data.

\bibliographystyle{elsarticle-num-names}
\bibliography{ref}

\appendix
\newpage
\appendix

\section{Numerical methods}
\label{app:numerical}
Here we describe the numerical methods used in this work and the implementation details.
Although the proposed MSR loss are generally applicable to any numerical method,
we use FDM as our numerical baselines in all experiments.
FDM is a class of numerical techniques for solving differential equations by approximating derivatives with finite differences.
By discretizing the spatial domain and time interval (if applicable) into a finite number of steps, 
FDM approximates differential operators with difference operators at each discrete point,
establishing an algebraic equation respect to the value of solution function at the point.
In this way, the value of solution function at these discrete points
can be obtained by solving the set of algebraic equations at all discrete points.
It is worth noting that boundary and initial value conditions can be naturally incorporated through constructing these algebraic equations.

In our experiments, we discretize the compact domain $\Omega\subseteq\mathbb{R}^2$
into a $n\times n$ uniform Cartesian grid with mesh width $\Delta$,
so that any function supported on $\Omega$ is approximated by its value on the grid points.
We denote the value of a function $u(x,y)$ at a gird point $(x_i, y_j)$ with $u_{i,j}$, 
where $x_i=x_0+i\Delta$ and $y_j=y_0+j\Delta$ for $i=0,\cdots,n-1$ and $j=0,\cdots,n-1$.

For the two-dimensional Poisson's equation shown in Eq.~(\ref{eq:poisson-preli}), we use second order central difference to approximate the Laplace operator, so we have
\begin{equation}\label{eq:discrete_poisson}
    \begin{aligned}
            \left(\frac{\partial^2 u}{\partial x^2} + \frac{\partial^2 u}{\partial y^2}\right)_{i,j} = &
            \frac{u_{i-1,j} - 2u_{i,j} + u_{i+1,j}}{\Delta^2} + \frac{u_{i,j-1} - 2u_{i,j} + u_{i,j+1}}{\Delta^2} \\
            = & \frac{u_{i,j-1} + u_{i-1, j} - 4u_{i,j} + u_{i,j+1} + u_{i+1,j}}{\Delta^2} = f_{i,j},
    \end{aligned}
\end{equation}
at each grid point $(x_i,y_j)$ for $i=1,\cdots,n-2$ and $j=1,\cdots,n-2$, 
and $u_{0,j}=u_{n-1,j}=0$ for $j=0,\cdots,n-1$ and $u_{i,0}=u_{i,n-1}=0$ for $i=0,\cdots,n-1$,
since we consider a zero Dirichlet boundary condition.
In this way, we have $(n-2)\times (n-2)$ linear algebraic equations for $(n-2)\times (n-2)$ unknown variables $u_{i,j}$, which can be solved with iterative methods.

For the two-dimensional Navier-Stokes equations shown in Eq.~(\ref{eq:nse}), in which vorticity function $\omega$ and steam function $\psi$ are time-dependent,
we discrete the time interval $[0,T]$ with time step $\Delta t$, 
and denote a discrete function $u$ at a gird point $(i,j)$ and $t$-th time step with $u_{i,j}^t$.
We consider a lid-driven cavity boundary condition, that is,
except for the top boundary where there is a flow with constant speed along the boundary,
other three boundaries are stationary walls, formally,
\begin{equation}\label{eq:lid-driven}
    \centering
    \begin{aligned}
        & \psi_{i,j}=0 \quad \text{for} \; i=0,n-1 \; \text{and} \; j=0,n-1 \\
        & \omega_{i,0}=-\frac{2}{\Delta^2}\psi_{i,1} \quad \text{for} \; i=0,\cdots,n-1 \\
        & \omega_{i,n-1} = -\frac{2}{\Delta^2}\psi_{i,n-2}-\frac{2}{\Delta}U \quad \text{for} \; i=0,\cdots,n-1 \\
        & \omega_{0,j}= -\frac{2}{\Delta^2}\psi_{1,j} \quad \text{for} \; j=0,\cdots,n-1 \\
        & \omega_{n-1,j}=-\frac{2}{\Delta^2}\psi_{n-2,j} \quad \text{for} \; j=0.\cdots,n-1 \\
    \end{aligned}
\end{equation}
where $U$ is the speed of constant flow.
Given $\omega^t_{i,j}$ for $i=1,\cdots,n-2$ and $j=1,\cdots,n-2$,
we first solve the Poisson's equation shown in Eq.~(\ref{eq:nse1}) based on Eq.~(\ref{eq:discrete_poisson}) and the first equation of Eq.~(\ref{eq:lid-driven}) to get $\psi^t_{i,j}$ for $i=1,\cdots,n-2$ and $j=1,\cdots,n-2$,
and then integrate Eq.~(\ref{eq:nse}) with the forward Euler method and Eq.~(\ref{eq:lid-driven}), that is,
\begin{equation}\label{eq:forward_euler}
\begin{aligned}
    \omega^{t+1}_{i,j} = \omega^t_{i,j} - & \frac{\Delta t}{4\Delta^2}\left[
    \left(\psi_{i,j+1}^t-\psi_{i,j-1}^t\right)\left(\omega_{i+1,j}^t-\omega_{i-1,j}^t\right) -
    \left(\psi_{i+1,j}^t-\psi_{i-1,j}^t\right)\left(\omega_{i,j+1}^t-\omega_{i,j-1}^t\right)
    \right] \\
    + & \frac{\Delta t}{\text{Re}\Delta^2}\left(
    \omega_{i+1,j}^t+\omega_{i-1,j}^t+\omega_{i,j+1}^t+\omega_{i,j-1}^t-4\omega_{i,j}^t\right),
\end{aligned}
\end{equation}
to obtain $w_{i,j}^{t+1}$ for $i=1,\cdots,n-2$ and $j=1,\cdots,n-2$.
In this way, given a initial condition $\omega_{i,j}^0$,
we can obtain $\omega_{i,j}^t$ by iteratively performing Eq.~(\ref{eq:discrete_poisson}) and Eq.~(\ref{eq:forward_euler}) $t$ times.

Finally, we use the conjugate gradient method to solve the systems of linear equations constructed in this work,
as they are all positive define. The conjugate gradient method is often implemented as an iterative algorithm, applicable to sparse systems that are too large to be handled by a direct implementation or other direct methods such as the Cholesky decomposition.
We run all FDM solvers described in this section with a V100 GPU.

\section{Experiment details} 
\label{app:exp_detail}
In this section, we will introduce the details on the configuration of the PDEs and the settings of training.
In all 2D experiments, we use the same difference scheme as the numerical methods introduced above. While for 3D, we use the staggered Marker-And-Cell (MAC) method. We conducted the experiments on uniform grids. The test data used for errors computation were generated using the numerical finite-difference method at the same discretization as the training data, though they were not exposed to the model during training. 

\subsection{Poisson's equation}

The functional parameter $f$ is generated from a random field satisfying the distribution $\mathcal{N}\left(0,7^{3/2}(-\Delta+49 I)^{-2.5}\right)$, where $I$ represents the identity matrix. 
Because the boundary of $u$ is fixed to zero, the neural network output $\widehat{u}$ does not include the boundary, which is a 2-dimensional matrix of the shape $(n-1)\times (n-1)$ where $n$ is the resolution.

Considering the linearity of the mapping from the network input $\widehat{f}$ to the output $\widehat{u}$, we construct LordNet without introducing any nonlinearities like activation functions. 
Specifically, in the case of the resolution $n=32$, we first add a $1\times 1$ Convolutional layer to transform the input to a feature map with 16 channels. Then we stack 2 multi-channel linear transformation, both of whom has 16 channels. At last, we add a $1\times 1$ Convolutional layer to reduce all channels into the one channel output.
In the case of the resolution $n=128$, we increase the channel count to 64, and the multi-channel linear transformation layers stacked to 4.
In addition, to increase the capacity of the neural network, we add a 1$\times$1 Convolutional layer between every two multi-channel linear transformation layers.
For comparison, we construct a linear Convolutional neural networks with the global dependency by stacking multiple Convolutional layers until the receptive field at any point of the mesh covers all other points. 
Because the receptive field of vanilla Convolutional layers grows linearly according to the layer count, we use the dilated Convolutional layers with the dilation steps growing 2 times each layer.

In all experiments, we set the initial learning rate as 1e-3 and decay the learning rate with a factor of 0.8 every 10,000 batches. The training batch size is 256, and the maximum iteration is 150,000.

\subsection{Navier-Stokes equation}
We consider the parametric setting where different initial states $\omega_0$ are sampled from a random field, satisfying the distribution $\mathcal{N}\left(0,8^{3}(-\Delta+64 I)^{-4.0}\right)$, the speed of constant flow $U$ is 1, and Reynolds number $\text{Re}$ is fixed to 1,000. 
For efficiency issues, the random field we use generates a periodic function, which, however, does not fit well to the lid-driven cavity boundary condition. For the periodic boundary condition, we generate the data on a 64$\times$64 grid with a time-step of 1e-2 where we record the solution every time step. For MSE training, we randomly sampled 6000 states in the whole time domain to train, while for the MSR loss, we only sampled initial states from random fields for training. In all experiments, we set the initial learning rate as 1e-3 and decay the learning rate with a factor of 0.9 every 50000 iterations. The training batch size is 64, and the maximum iteration is 500,000.

As for the lid-driven cavity, to get a smoother initial state, we solve with the numerical solver for the first $T_0$ seconds and use $\omega_{T_0}$ as the initial state, where $T_0=1.98$ in our experiments.
We set $\delta t = 0.01$, which is small enough to ensure the stability of the finite difference scheme and set the terminal time $T=30$ to ensure that the fluid becomes steady at the terminal state.
Therefore, there are 3,000 time steps from the initial state to the terminal state.
We collect 5,000 initial states For the training with MSR loss,  while for supervised training with MSE loss, we collect 5,000 states from $T_{start}=1.98$ to $T_{end}=30$.
To prepare the dataset for the MSE loss, we solve the PDE one timestep forward to get the next timestep, which is served as the labels.
For the test set, we fix $T_0=2$ to get a stable measurement on how a model performs. FDM is used to collect 25 trajectories with different initial states for testing, each of which contains 2,700 frames.
Similar to the case of Poisson's equation, we also train the neural network to predict the values of $\widehat{\psi}$ inside the boundary, which is a 2-dimensional matrix of the shape $(n-2)  \times (n-2)$ where $n$ is the resolution.

We implement ResNet with the ResNet blocks based on the official implementation of PyTorch~\citep{paszke2017automatic}. It contains 10 ResNet blocks with the channel counts ranging from 16 to 128. 
In addition, we use the LayerNorm~\citep{ba2016layer} in all normalization layers, because it performs better than BatchNorm~\citep{ioffe2015batch}.
For Swin Transformer, we migrate the official implementations with the default settings to our project.
For FNO, we use the 2-dimensional version in the official repository.
For the proposed LordNet, we only stack 2 Lord modules and fix the channel count to 64 in all layers. 
In the position-wise embedding of the 2 Lord modules, we stack 2 1$\times$1 Convolutional layers, where the hidden embedding contains 256 and 128 channels, respectively, and GELU activation is used between the Convolutional layers.

In experiments with the same neural network, we use the same hyperparameter for training with MSR loss and MSE loss.
In addition, we set the initial learning rate as 1e-3 and decay the learning rate with a factor of 0.9 every 100 epochs in all experiments. The training batch size is 64, and the maximum epochs is 5,000.

\subsection{3D fluid problem}
\label{app:3d}
We generate training data with code in \cite{wandel_teaching_2021}, where the resolution of the domain is 128$\times$64$\times$64 and $\delta t=4$. 
Various shapes of obstacles are used during the training, including boxes, spinning balls, or cylinders. The locations of the obstacles vary from 22 to 42 for y and z coordinates and 86 to 106 for x coordinates, and the diameter of the obstacle varies from 10 to 54. The maximum inflow/outflow velocity is 3.0 m/s.
For all experiments, we use a learning rate of 1e-3 and Adam optimizer to train the model for 1000 epochs with a batch size of 10. To conduct a fair comparison, we use the same configuration of the training data generation.
The U-net3d (baseline) is the trained model from the repository of \cite{wandel_teaching_2021}. For the FNO3d, we set the truncation mode to 12 and the width to 64. For the LordNet3d, we only stack 2 Lord modules and fix the channel count to 64 in all layers. The position-wise embedding contains two 1$\times$1 convolutional layers whose hidden embedding channels are 256 and 128, respectively. The activation function for FNO3d and LordNet3d is GELU activation.
The quantitative comparison is based on the PDE residuals on the 128$\times$64$\times$64 domain.
It should be noted that no trajectories generated from the numerical simulation are incorporated as the input during either the training or inference stages. The snapshots shown in the paper are the predictions after doing 116 autoregressive inferences using trained neural networks.

\end{document}